\documentclass[letterpaper]{article} 

\usepackage{iclr2026_conference,times}

\usepackage{amsmath,amsfonts,bm}

\def\eqref#1{equation~\ref{#1}}

\def\1{\bm{1}}

\DeclareMathAlphabet{\mathsfit}{\encodingdefault}{\sfdefault}{m}{sl}
\SetMathAlphabet{\mathsfit}{bold}{\encodingdefault}{\sfdefault}{bx}{n}

\usepackage{hyperref}
\usepackage{url}

\usepackage{graphicx}
\usepackage{booktabs}
\usepackage{amsmath}
\usepackage{amssymb}
\usepackage{mathtools}
\usepackage{amsthm}
\usepackage[capitalize,noabbrev]{cleveref}

\usepackage{pgfplots}
\usepackage{pgfplotstable}
\pgfplotsset{compat=1.18}
\usetikzlibrary{arrows.meta,positioning,shapes.geometric,shapes.misc,fit,calc,shadows.blur}

\newcommand{\hflogo}{\raisebox{-0.25\height}{\includegraphics[width=1.26em]{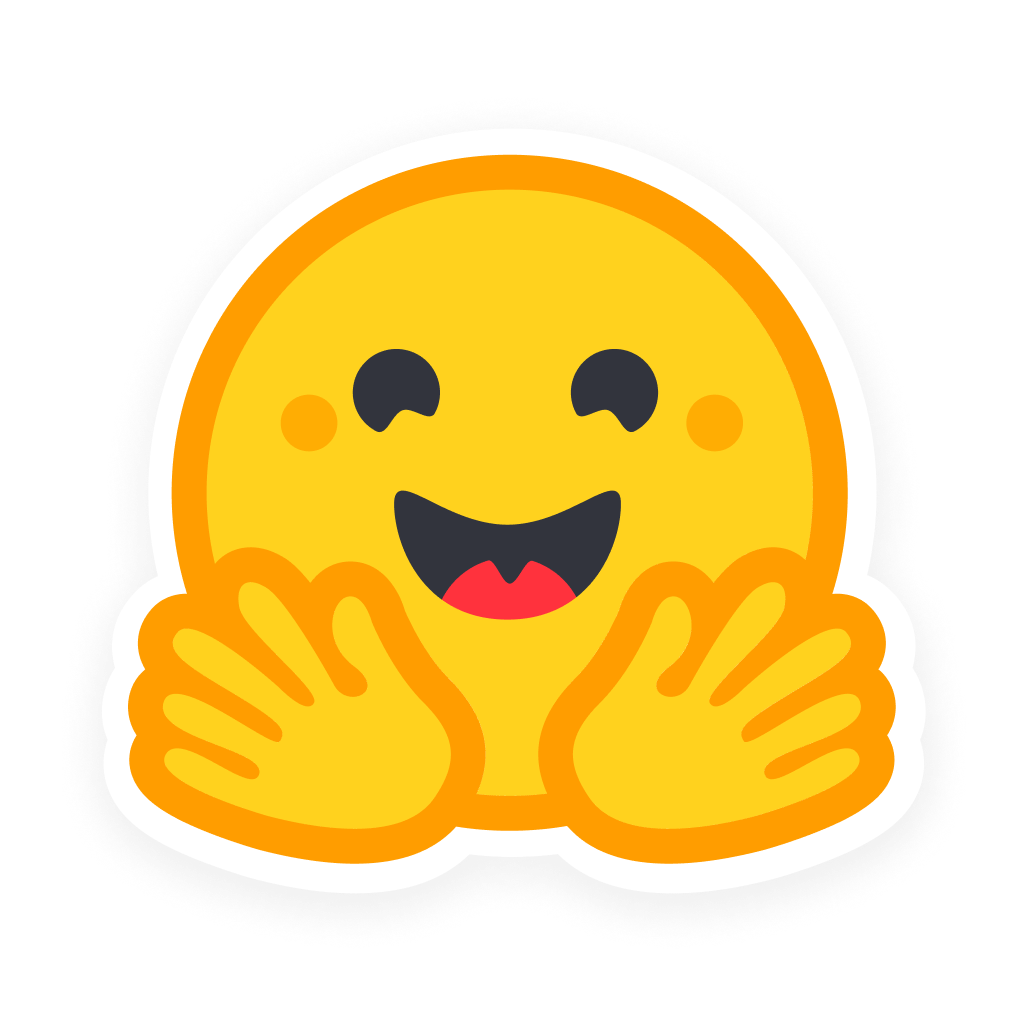}}}
\newcommand{\matslogo}{\raisebox{-0.25\height}{\includegraphics[width=0.45em]{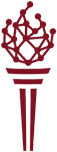}}}
\newcommand{\sparlogo}{\raisebox{-0.25\height}{\includegraphics[width=0.9em]{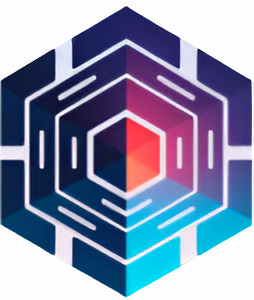}}}

\theoremstyle{plain}

\theoremstyle{definition}

\theoremstyle{remark}

\title{Expert Selections In MoE Models Reveal \\ (Almost) As Much As Text}

\author{%
Amir Nuriyev \\
MBZUAI \\
\And
Gabriel Kulp \\
RAND, Oregon State University
}

\iclrfinalcopy

\begin{document}
\maketitle

\begin{abstract}
We present a text-reconstruction attack on mixture-of-experts (MoE) language models that recovers tokens from expert selections alone. In MoE models, each token is routed to a subset of expert subnetworks; we show these routing decisions leak substantially more information than previously understood. Prior work using logistic regression achieves limited reconstruction; we show that a 3-layer MLP improves this to 63.1\% top-1 accuracy, and that a transformer-based sequence decoder recovers 91.2\% of tokens top-1 (94.8\% top-10) on 32-token sequences from OpenWebText after training on 100M tokens. These results connect MoE routing to the broader literature on embedding inversion. We outline practical leakage scenarios (e.g., distributed inference and side channels) and show that adding noise reduces but does not eliminate reconstruction. Our findings suggest that expert selections in MoE deployments should be treated as sensitive as the underlying text.\footnote{Our dataset and code are available on \href{https://huggingface.co/datasets/anpaurehf/openwebexpertselections}{\hflogo\ HuggingFace}.}\footnote{This work was conducted while Gabriel Kulp was a graduate student at Oregon State University. He is currently affiliated with RAND as an adjunct Technology and Security Policy Fellow (see \href{https://www.rand.org/global-and-emerging-risks/centers/ai-security-and-technology/fellows.html}{www.rand.org/cast/fellows} for more information).}
\end{abstract}

\section{Introduction}
As modern large language models have grown in scale, there is growing demand for compute-efficient transformer architectures. Mixture-of-experts (MoE) models address this by activating only a subset of parameters per token, speeding up training and inference \citep{shazeer2017outrageously,fedus2022switch,jiang2024mixtral}. Consequently, MoE architectures are widely used in modern LLMs. This motivates an examination of the expert-selection mechanism in these models. In this work, we show an attack that exploits MoE routing: the expert selections can leak enough information to reconstruct the underlying text.

\section{Related Work}
\paragraph{Embedding inversion.}
A closely related line of work studies inversion of continuous text representations. \citet{morris2023text} propose \emph{vec2text}, showing that a learned decoder can reconstruct text from sentence embeddings, demonstrating that embedding vectors can leak substantial lexical and semantic content; see also earlier analyses of embedding leakage \citep{song2020leakage}. \citet{zhang2025zsinvert} propose \emph{ZS-Invert}, which targets black-box embedding APIs and performs universal, zero-shot inversion without training an embedding-specific decoder. \citet{huang2024transferable} study transferable embedding inversion, showing that an attacker can recover text from embeddings without querying the target embedding model.

Our setting is different: expert selections are a \emph{discrete} and lower-bandwidth intermediate signal than full embedding vectors or hidden states, but are emitted repeatedly across layers and tokens. We show that even these expert-selection traces can support high-fidelity reconstruction.

\paragraph{MoE side channels.}
\citet{ding2025moecho} show that MoE routing information can leak via architectural side channels (e.g., GPU performance counters), and demonstrate downstream prompt inference and response reconstruction using a logistic regression decoder and templated-prompt strategies. Our work complements this: we study the decoding problem given (possibly partial) expert-selection traces and show that sequence-level decoding can substantially improve reconstruction over per-token classifiers. As a stronger baseline than logistic regression, we use a 3-layer MLP decoder. We also describe additional leakage surfaces beyond the specific side-channel instantiations considered in MoEcho, such as distributed inference and pipeline-parallel MoE.

\paragraph{Model theft and extraction.}
Broader work on attacks against deployed language models studies extracting sensitive information or model components from production systems \citep{carlini2024stealing}, complementing our focus on privacy leakage from intermediate routing signals.

\paragraph{Output inversion and prompt extraction.}
Related work also studies reconstructing prompts from observable model outputs. \citet{zhang2024extracting} extract prompts by inverting LLM outputs, which is complementary to our setting where the observed signal is an internal routing trace rather than generated text.

\section{Threat Model}
\label{sec:threat_model}
\textbf{Observed signal.} The adversary observes only the router's expert selections for each token at one or more layers. The adversary may observe a subset of layers and does not observe router logits, router weights, hidden states, or expert outputs.

\textbf{Goal.} Given the set of selected experts corresponding to an unknown token sequence, the adversary aims to reconstruct the underlying text, ideally recovering a semantically similar sequence if not the exact tokens.

\textbf{Auxiliary knowledge at attack time.} We assume the adversary knows the tokenizer used by the victim model and the MoE routing configuration (e.g., the number of experts and $k$). In our experiments, we assume the model family and routing configuration match \texttt{gpt-oss-20b}.

\textbf{Data access for learning-based decoders.} For our MLP and sequence decoder, we assume the adversary can obtain training pairs of ``(token, expert-selection trace)'' from a same-family model, or from other sources that expose both text and expert-selection traces (e.g., internal logs in distributed inference). We trained the sequence decoder on contiguous token sequences (not shuffled tokens), preserving the natural order in which tokens (and therefore selected experts) appear.

\section{Attack Surfaces}
We now describe practical settings in which an adversary may obtain expert-selection traces. The key observation is that expert selections are a low-throughput signal, but they can be exposed whenever routing decisions cross boundaries (e.g., between devices, processes, or administrative domains) or leak through side channels. An adversary can collect ``(text, trace)'' pairs from benign workloads where text is known, train an inverter on these pairs, then apply the trained inverter to reconstruct text from traces of sensitive workloads.

\paragraph{Distributed inference.}
In distributed inference settings, a malicious host machine running a model (or a subset of its layers or experts) can observe full or partial routing traces and decode the original text, violating user privacy.

\paragraph{Physical side channels.}
In supply-chain or co-residency scenarios, attackers may be able to collect side channel measurements (e.g., via power draw or electromagnetic emissions) to infer which experts are selected at runtime, then map these routing traces to confidential tokens using our decoding method. Our experiments in this paper are focused on the latter half of this attack, assuming the adversary has already identified routing traces; prior work demonstrates that MoE routing can be inferred via architectural side channels (e.g., GPU performance counters) \citep{ding2025moecho}, suggesting that physical side channels are a plausible additional leakage route.

\paragraph{Pipeline-parallel MoE.}
If experts are sharded across data center nodes (common in pipeline-parallel MoE), an adversary might only need to detect which GPU exhibits activity over time, then infer the responsible experts via frequency analysis (or get expert identification for free when each expert uniquely maps to a single device).

\section{Decoding Attack}
We use OpenWebText for evaluation and training since it contains a wide variety of text, including high-entropy data like passwords and API keys. To obtain training data, we run \texttt{gpt-oss-20b} (32 experts, top-4 routing, 24 layers, vocab size 201{,}088) over 100M tokens of OpenWebText split into 32-token chunks in prefill (no autoregressive generation), yielding ``(token sequence, expert-selection trace)'' pairs. Here an \emph{expert-selection trace} (also called a \emph{routing trace}) consists of the router's unordered top-$k$ expert indices for each token at each observed layer. We use these terms interchangeably throughout. Top-1, top-5, and top-10 exact token decoding accuracies are evaluated on a held-out OpenWebText split (10M tokens) disjoint from training; unless otherwise stated, the trace includes expert selections from all 24 layers.

\paragraph{Model and notation.}
Let $x_{1:T} = (x_1,\ldots,x_T)$ denote a token sequence of length $T$. For a model with $L$ MoE layers, $n$ experts per layer, and top-$k$ routing, let $\phi$ be the (deterministic) routing trace function mapping a token sequence to the per-layer expert selections.
\begin{align}
  I &= \phi(x_{1:T}), \\
  I &= \left(I_{\ell,t}\right)_{\ell=1,\,t=1}^{L,\,T}, \\
  I_{\ell,t} &\subseteq \{1,\ldots,n\}, \quad |I_{\ell,t}|=k.
\end{align}
Here $I_{\ell,t}$ is the \emph{unordered} set of the $k$ experts selected for token $x_t$ at layer $\ell$. The attacker observes $I$ and aims to recover $x_{1:T}$.

For learning-based inversion, we train a decoder $p_\theta(x_{1:T}\mid I)$ by maximum likelihood, i.e., minimizing the negative log-likelihood over training pairs $(x_{1:T}, I)$.

\begin{figure}[t]
  \centering
  \begin{tikzpicture}
    \begin{axis}[
      ybar,
      bar width=14pt,
      width=0.9\columnwidth,
      height=0.45\columnwidth,
      ymin=0, ymax=100,
      ylabel={Accuracy (\%)},
      symbolic x coords={3-layer MLP, Sequence decoder},
      xtick=data,
      enlarge x limits=0.30,
      ymajorgrids,
      grid style={gray!15},
      tick label style={font=\scriptsize},
      label style={font=\scriptsize},
      legend style={
        at={(0.5,1.02)},
        anchor=south,
        draw=none,
        fill=none,
        font=\scriptsize,
        legend columns=3,
        /tikz/every even column/.append style={column sep=0.8em},
      },
      legend image code/.code={\draw[#1, fill opacity=0.7] (0pt,-1pt) rectangle (6pt,5pt);},
      every axis plot/.append style={draw opacity=1, fill opacity=0.55, line width=0.35pt},
      nodes near coords,
      nodes near coords style={font=\tiny, text=black, anchor=north, yshift=-1.5pt},
      point meta=y,
      nodes near coords align={vertical},
    ]
      \addplot[draw=blue!70!black, fill=blue!35]  coordinates {(3-layer MLP,63.1) (Sequence decoder,91.2)};
      \addplot[draw=red!70!black,  fill=red!25]   coordinates {(3-layer MLP,80.3) (Sequence decoder,94.3)};
      \addplot[draw=brown!70!black,fill=brown!25] coordinates {(3-layer MLP,84.3) (Sequence decoder,94.8)};
      \legend{Top-1, Top-5, Top-10}
    \end{axis}
  \end{tikzpicture}
  \caption{Accuracy of decoding tokens from expert selections on OpenWebText.}
  \label{fig:decoder_accuracy}
\end{figure}
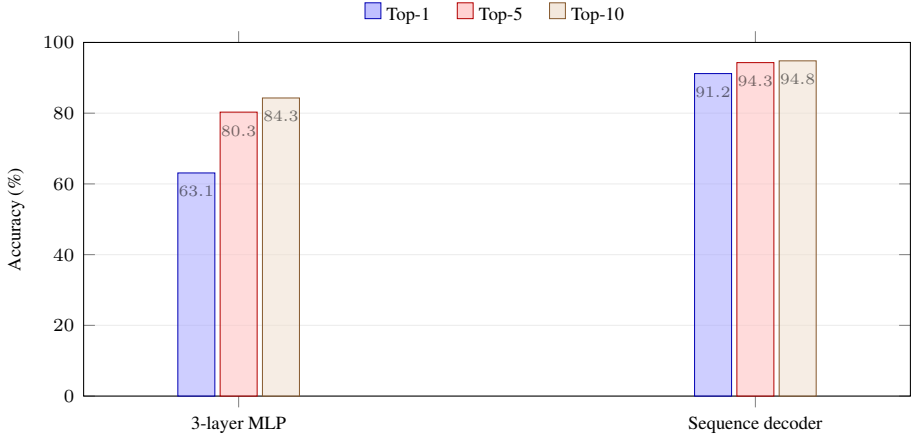

\paragraph{Single-token MLP.}
A 3-layer MLP trained to predict a token from its expert selections obtains 63.1\% top-1 accuracy (80.3\% top-5, 84.3\% top-10; \Cref{fig:decoder_accuracy}). This decoder treats each token independently, learning a mapping from a single token's expert-selection trace to a distribution over the vocabulary. Ablations indicate that performance declines when the MLP has more than six layers.

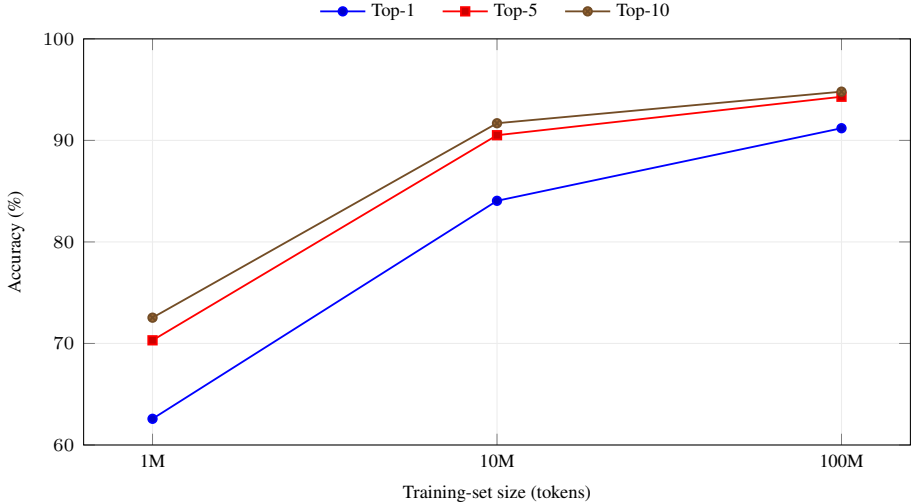
\begin{figure}[t]
  \centering
  \begin{tikzpicture}
    \begin{axis}[
      width=0.9\columnwidth,
      height=0.50\columnwidth,
      ymin=60, ymax=100,
      xlabel={Training-set size (tokens)},
      ylabel={Accuracy (\%)},
      symbolic x coords={1M, 10M, 100M},
      xtick=data,
      ymajorgrids,
      xmajorgrids,
      grid style={gray!15},
      tick label style={font=\scriptsize},
      label style={font=\scriptsize},
      legend style={
        at={(0.5,1.02)},
        anchor=south,
        draw=none,
        fill=none,
        font=\scriptsize,
        legend columns=3,
        /tikz/every even column/.append style={column sep=0.8em},
      },
      every axis plot/.append style={line width=0.7pt, mark=*, mark size=1.6pt},
    ]
      \addplot coordinates {(1M,62.58) (10M,84.05) (100M,91.20)};
      \addplot coordinates {(1M,70.31) (10M,90.51) (100M,94.30)};
      \addplot coordinates {(1M,72.54) (10M,91.69) (100M,94.80)};
      \legend{Top-1, Top-5, Top-10}
    \end{axis}
  \end{tikzpicture}
  \caption{Accuracy vs. training-set size for the sequence decoder on OpenWebText.}
  \label{fig:acc_vs_size}
\end{figure}

\paragraph{Sequence decoder.}
We train an encoder-only transformer that maps the expert-selection trace to the token sequence. Unlike the MLP baseline, the transformer consumes the entire length-$T$ expert-selection trace jointly and predicts the full token sequence, allowing it to exploit dependencies across positions. Our implementation first converts the per-layer top-$k$ expert selections into 32-dimensional binary vectors with exactly 4 ones (to represent \texttt{gpt-oss-20b} with $n=32, k=4$), applies a small per-layer MLP and concatenates representations across the layers of the MoE model under observation, then projects into a token-level embedding stream with learned positional embeddings. We then apply a stack of non-causal self-attention blocks and predict token logits with a linear head, training with cross-entropy on the observed positions. This model achieves 91.2\% top-1 accuracy (94.3\% top-5, 94.8\% top-10) on 10M held-out OpenWebText tokens when trained on traces from 100M tokens, substantially outperforming the MLP baseline. We observe that accuracy degrades gracefully with less training data (\Cref{fig:acc_vs_size}); accuracy as a function of token frequency is shown in \Cref{fig:freq_vs_acc}.

\section{Information Leakage From Expert Selections}
Router outputs for a token are conditioned on the context and therefore deterministic for a fixed prefix, up to floating-point error. In practice, near-ties in top-$k$, GPU nondeterminism, and quantization can cause occasional flips. Recent work argues that hidden states in decoder-only transformers are almost surely injective and hence invertible \citep{nikolaou2025injective}. One might conjecture that router logits (as smooth images of these states) preserve token identity with high probability. However, quantization and floating-point errors can break strict invertibility, suggesting expert selections might not allow exact reconstruction.

Despite that, we empirically show that expert selections still provide ample information for decoding, while relaxing threat-model assumptions and enabling additional attack surfaces. For example, side-channel approaches such as MoEcho \citep{ding2025moecho} use NVIDIA Performance Counters to infer activated experts and decode those inferred expert selection traces into text using a logistic regression decoder. Conceptually, expert selections resemble discrete ``embeddings'' of tokens and contexts, connecting our setting to existing embedding inversion attacks \citep{morris2023text}.

We define \emph{expert selections} at a layer as the unordered set of experts selected by top-$k$ routing among $n$ experts. Let $I_\ell$ denote this set at layer $\ell$. For a single layer, the number of possible selections is $\binom{n}{k}$, so the entropy is bounded by
\begin{equation}
  H(I_\ell) \le \log_2 \binom{n}{k} \quad \text{bits,}
\end{equation}
with equality only if selections are uniform over all $\binom{n}{k}$ outcomes.

Across $L$ layers, let $I_{1:L}=(I_1,\ldots,I_L)$ be the routing trace for a token. By subadditivity of entropy,
\begin{equation}
  H(I_{1:L}) \le \sum_{\ell=1}^L H(I_\ell) \le L\,\log_2 \binom{n}{k}.
\end{equation}
For \texttt{gpt-oss-20b} ($n=32, k=4$) with $L=24$, this yields an (extremely loose) upper bound of $24\log_2\binom{32}{4} \approx 363$ bits per token.

This value should not be read as ``bits of token identity.'' In practice, selections are correlated across layers and depend on context, so the effective entropy is substantially lower.

\section{Layerwise Information Analysis}
Which layers are most informative? We estimate the layerwise entropy of expert selections and the mutual information between layers. \Cref{fig:entropy_by_layer} visualizes the resulting entropy profile across the 24 layers, computed with plug-in estimators over empirical selection distributions.

\paragraph{Random variables.}
Let $t$ be a uniformly random token position in our trace dataset. For each MoE layer $\ell$, define the routing random variable
\begin{equation}
  I_\ell := I_{\ell,t},
\end{equation}
where $I_{\ell,t}$ is the \emph{unordered} set of the top-$k$ expert selections for token position $t$ at layer $\ell$.

\paragraph{Per-layer entropy estimation.}
Let $\hat{p}_\ell(S)$ be the fraction of token positions whose selection at layer $\ell$ equals expert set $S$. We estimate the entropy with the plug-in estimator
\begin{equation}
\widehat{H}(I_\ell) \;=\; - \sum_{S \in \mathcal{S}_\ell} \hat{p}_\ell(S)\,\log_2 \hat{p}_\ell(S),
\end{equation}
where $\mathcal{S}_\ell$ is the set of distinct expert sets observed at least once at layer $\ell$ (unobserved outcomes are treated as having probability 0). Given the large sample size, we expect finite-sample bias to be small for $\widehat{H}(I_\ell)$.

\begin{figure}[t]
  \centering
  \begin{tikzpicture}
    \pgfplotstableread[col sep=comma]{
layer,entropy_bits,entropy_per_expert,support_size
0,9.517046697705933,2.379261674426483,5742
1,9.487699655467605,2.371924913866901,4308
2,8.976020916722394,2.2440052291805985,3743
3,8.65239232368745,2.1630980809218623,3815
4,9.004515434024787,2.251128858506197,4676
5,9.148341193406607,2.2870852983516516,5008
6,9.902790981816604,2.475697745454151,5872
7,9.200958217465232,2.300239554366308,4012
8,8.564088451389777,2.1410221128474443,3258
9,8.348805356023458,2.0872013390058646,3336
10,7.353785350196574,1.8384463375491435,2278
11,7.949635456005803,1.9874088640014507,2210
12,8.25661392676017,2.0641534816900426,2580
13,8.148422500834323,2.037105625208581,2368
14,8.472944261690513,2.118236065422628,2777
15,8.863288425256586,2.2158221063141466,3418
16,8.535940274499463,2.133985068624866,2963
17,8.260001271433485,2.065000317858371,2841
18,7.995286991497832,1.998821747874458,2300
19,8.539388335839,2.13484708395975,3047
20,8.543022800909215,2.1357557002273038,3490
21,8.472601476903954,2.1181503692259884,3262
22,7.874538438303263,1.9686346095758158,2667
23,7.616667197716112,1.904166799429028,3401
    }\entropybylayer

    \begin{axis}[
      width=0.98\columnwidth,
      height=0.40\columnwidth,
      ymin=7.0,
      ymax=10.5,
      xmin=1,
      xmax=24,
      xlabel={Layer $\ell$},
      ylabel={$\widehat{H}(I_\ell)$ (bits)},
      xtick={1,2,3,4,5,6,7,8,9,10,11,12,13,14,15,16,17,18,19,20,21,22,23,24},
      xticklabel style={font=\tiny},
      tick label style={font=\scriptsize},
      label style={font=\scriptsize},
      ymajorgrids,
      grid style={gray!15},
      every axis plot/.append style={line width=0.9pt, mark=*, mark size=1.2pt, draw=blue!70!black},
    ]
      \addplot table[x expr=\thisrow{layer}+1,y=entropy_bits]{\entropybylayer};
    \end{axis}
  \end{tikzpicture}
  \caption{Estimated per-layer entropy of expert selections. The sum across all layers is 206 bits, which is an upper bound for the total router information content per forward pass.}
  \label{fig:entropy_by_layer}
\end{figure}

\paragraph{Mutual information heatmap construction.}
We compute plug-in estimates of inter-layer mutual information to characterize redundancy in routing patterns across layers; we use these estimates primarily for qualitative comparison. Early layers (1-7) show high mutual information with each other, while middle layers (particularly around layer 11) show reduced mutual information with both early and late layers, suggesting distinct routing regimes.

For each pair of layers $(i,j)$ with $i<j$, we accumulate empirical counts over observed pairs $(I_{i,t}, I_{j,t})$ across token positions, yielding an empirical joint distribution $\hat{p}_{ij}$ and marginals $\hat{p}_i, \hat{p}_j$. We then compute the plug-in mutual information estimator
\begin{equation}
\widehat{I}(I_i; I_j) \;=\; \sum_{(a,b)\in \mathcal{P}_{ij}} \hat{p}_{ij}(a,b)\,
\log_2\!\left(\frac{\hat{p}_{ij}(a,b)}{\hat{p}_i(a)\,\hat{p}_j(b)}\right),
\end{equation}
where $\mathcal{P}_{ij}$ ranges over expert-set pairs observed at least once (unobserved pairs are treated as having probability 0). 

\begin{figure}[t]
  \centering
  \begin{minipage}{0.50\columnwidth}
    \centering
    \begin{tikzpicture}
      \pgfplotstableread[col sep=comma]{
layer_i,layer_j,mutual_information_bits
0,1,6.793430485935312
0,2,6.141708689270875
0,3,5.860889902127956
0,4,5.727343197911402
0,5,5.988148579796868
0,6,6.317463475804468
0,7,5.956250221873838
0,8,5.310534683703818
0,9,4.994317340498276
0,10,4.053257956183871
0,11,4.371858791998377
0,12,4.579122825478291
0,13,4.468407300190313
0,14,5.091161600314489
0,15,5.409047861506696
0,16,5.0585213706755825
0,17,5.053750434324097
0,18,5.375638216410817
0,19,5.341106444669617
0,20,5.562015840221607
0,21,5.21643335697576
0,22,4.860251384127969
0,23,4.482691254129085
1,2,6.079556439907592
1,3,5.717813528287788
1,4,5.689457295818689
1,5,5.898810002095165
1,6,6.208846730549225
1,7,5.878131262031174
1,8,5.341150889671427
1,9,4.992283976208587
1,10,4.053461496967028
1,11,4.380117310349427
1,12,4.61616716165064
1,13,4.456542509551798
1,14,5.0902041945703385
1,15,5.370308266680959
1,16,4.9844595158829295
1,17,4.987036929211285
1,18,5.30768239849115
1,19,5.300747969865043
1,20,5.4733161035579565
1,21,5.082539478471683
1,22,4.7647469911193365
1,23,4.34920449880018
2,3,5.4754459643852
2,4,5.513173592687707
2,5,5.729604633759548
2,6,5.91997378438778
2,7,5.670964098426818
2,8,5.260330290236466
2,9,4.923123543956795
2,10,4.025627508819228
2,11,4.351572763318662
2,12,4.508494440141581
2,13,4.38514468438992
2,14,4.951500765358013
2,15,5.18293513981063
2,16,4.816087598528187
2,17,4.860935836289226
2,18,5.095302560311991
2,19,5.08487829376286
2,20,5.188040053366167
2,21,4.78705954323943
2,22,4.5152986043849195
2,23,4.032430217834415
3,4,5.282144198800635
3,5,5.475188773115028
3,6,5.563760983059373
3,7,5.2607013112814345
3,8,4.888271015632517
3,9,4.551007485766881
3,10,3.7728756424647547
3,11,4.098959941214281
3,12,4.221569604997724
3,13,4.134719613558653
3,14,4.5927015851522786
3,15,4.829304031274905
3,16,4.468668551256867
3,17,4.46274246832877
3,18,4.735767503546064
3,19,4.75080735365342
3,20,4.916745537521464
3,21,4.567527909499857
3,22,4.262163568077005
3,23,3.8656361921957805
4,5,5.599857694661368
4,6,5.8673310666375045
4,7,5.4663583959376245
4,8,5.074077887710603
4,9,4.818690845511421
4,10,4.0242549070764495
4,11,4.343217330648364
4,12,4.494770502809212
4,13,4.3566004881566505
4,14,4.771689743146557
4,15,4.957603940868831
4,16,4.618045951031328
4,17,4.605596419517569
4,18,4.750554754278428
4,19,4.886143995509363
4,20,4.931221611679898
4,21,4.65105751100336
4,22,4.279332418553449
4,23,3.9023190447652008
5,6,5.997302230639159
5,7,5.657066920773726
5,8,5.2671992693521235
5,9,5.001605369242813
5,10,4.190757099802981
5,11,4.5251585019709335
5,12,4.655750477454501
5,13,4.509963402980496
5,14,4.9652421083928875
5,15,5.198988529574815
5,16,4.79917734179237
5,17,4.884528013790206
5,18,5.052317736000925
5,19,5.130751795674238
5,20,5.234217357884045
5,21,4.875720974566925
5,22,4.534678048926221
5,23,4.091098876890765
6,7,6.179707098295094
6,8,5.729787279251223
6,9,5.267344511741964
6,10,4.375065544507972
6,11,4.716027097452932
6,12,4.9084900733633186
6,13,4.761567834892441
6,14,5.319757283661587
6,15,5.51287682467325
6,16,5.205279978501642
6,17,5.09867926629549
6,18,5.265099631245414
6,19,5.326808641971337
6,20,5.456716433879222
6,21,5.134238338042907
6,22,4.759386889831837
6,23,4.385046234075509
7,8,5.666679479092991
7,9,5.217019177816568
7,10,4.290702868933372
7,11,4.589802317933157
7,12,4.746040798752302
7,13,4.619545114586977
7,14,5.272238464413399
7,15,5.4803402678999396
7,16,5.125893349701982
7,17,4.992341361183804
7,18,5.195857821306304
7,19,5.188926142953137
7,20,5.271175629908377
7,21,4.907223092919501
7,22,4.568916613212904
7,23,4.18624162009564
8,9,5.041862482272608
8,10,4.196422124318251
8,11,4.471577417143038
8,12,4.605571980346856
8,13,4.458520276978312
8,14,5.161310544536023
8,15,5.246449737528928
8,16,4.966899019281636
8,17,4.77176023263311
8,18,4.87589732311212
8,19,4.886382630467513
8,20,4.937246454709756
8,21,4.540587718237046
8,22,4.268777638793386
8,23,3.860203467603734
9,10,4.293651005666346
9,11,4.550685739907165
9,12,4.628681224720222
9,13,4.538131820829846
9,14,4.915384596621803
9,15,4.9642893596899835
9,16,4.64883105080499
9,17,4.559117462393754
9,18,4.546950849350309
9,19,4.617283489295565
9,20,4.6516285955789725
9,21,4.289866279263504
9,22,3.9951537755924615
9,23,3.6467088181292087
10,11,4.138399660085222
10,12,4.275629147841065
10,13,4.1436189985709175
10,14,4.122236098739671
10,15,4.113920869881897
10,16,3.8127074189817964
10,17,3.7843547057771025
10,18,3.663264667679217
10,19,3.8211680006808377
10,20,3.791850432118036
10,21,3.509113250854656
10,22,3.236729450253792
10,23,3.006664615067322
11,12,4.492608199233904
11,13,4.418847670939734
11,14,4.464577424377641
11,15,4.442718227149569
11,16,4.175983833048301
11,17,4.1257600045642535
11,18,4.016330539746071
11,19,4.184788374415651
11,20,4.165466282959455
11,21,3.8851214221759487
11,22,3.5831970071073815
11,23,3.3089052078161707
12,13,4.621591012200828
12,14,4.636988311020005
12,15,4.655875486395938
12,16,4.34269974209963
12,17,4.301612106778576
12,18,4.202770732354616
12,19,4.373916049502419
12,20,4.335248045068559
12,21,4.054472943074172
12,22,3.7400370762825967
12,23,3.490367938650306
13,14,4.5429257601125625
13,15,4.524692609188254
13,16,4.269907394732724
13,17,4.211946425910757
13,18,4.090491034170406
13,19,4.26031764456513
13,20,4.225644556036326
13,21,3.9544506012893366
13,22,3.6550865313984406
13,23,3.3783567244950787
14,15,5.484381907989414
14,16,5.150617760852899
14,17,4.850025043471258
14,18,4.934556379821606
14,19,4.8906101738698755
14,20,4.914046510790717
14,21,4.485970741508448
14,22,4.240811918987311
14,23,3.8173989981340473
15,16,5.379766573912863
15,17,5.095307364467514
15,18,5.2233921782792505
15,19,5.168116755791658
15,20,5.228559214043499
15,21,4.794316373317175
15,22,4.510958076142805
15,23,4.041203440402476
16,17,5.047518386959993
16,18,5.049252965044237
16,19,5.0559940413130775
16,20,5.091861715442086
16,21,4.714705628489388
16,22,4.4637199136016505
16,23,4.085316738653629
17,18,5.046954016835593
17,19,5.213983981247108
17,20,5.215227969772547
17,21,4.839130240929037
17,22,4.580574725623421
17,23,4.1365277314069795
18,19,5.33893442483435
18,20,5.47760117612711
18,21,4.997720529249087
18,22,4.764143149570301
18,23,4.1647618854537605
19,20,5.623295014808992
19,21,5.264408461435253
19,22,4.978274167405652
19,23,4.460894804408856
20,21,5.486651390088862
20,22,5.280779928644999
20,23,4.730476216464224
21,22,5.041605563846866
21,23,4.632795000767333
22,23,4.539143464588938
      }\mipairs

      \begin{axis}[
        width=\linewidth,
        height=\linewidth,
        xmin=0.5,
        xmax=24.5,
        ymin=0.5,
        ymax=24.5,
        xlabel={Layer $i$},
        ylabel={Layer $j$},
        xtick={1,9,17,24},
        ytick={1,9,17,24},
        axis on top,
        enlargelimits=false,
        axis equal image,
        colorbar,
        colormap/viridis,
        point meta min=0,
        point meta max=6.9,
        tick label style={font=\scriptsize},
        label style={font=\scriptsize},
        colorbar style={
          tick label style={font=\scriptsize},
          ylabel={Mutual information (bits)},
          ylabel style={font=\scriptsize},
        },
      ]
        \addplot[
          scatter,
          only marks,
          point meta=explicit,
          scatter/use mapped color={draw opacity=0, fill=mapped color},
          mark=square*,
          mark size=3.3pt,
        ] table[
          x expr=\thisrow{layer_j}+1,
          y expr=\thisrow{layer_i}+1,
          meta=mutual_information_bits,
          restrict expr to domain={\thisrow{layer_i}-\thisrow{layer_j}}{-1000:-1}
        ]{\mipairs};

      \end{axis}
    \end{tikzpicture}
  \end{minipage}
  \caption{Estimated mutual information between layers' expert selections.}
  \label{fig:mi_heatmap}
\end{figure}

\section{Mitigations}
Because our attack assumes access to expert selections, the most direct mitigation is to treat expert-selection traces as sensitive outputs and minimize their exposure. Concretely, production deployments should avoid returning, logging, or exporting per-token expert selections unless also handling tokens the same way. Production deployments should treat routing information as though it is the same as tokens, especially when such information crosses trust boundaries between tenants, machines, or administrative domains. 

\begin{figure}[t]
  \centering
  \begin{tikzpicture}
    \begin{axis}[
      width=0.9\columnwidth,
      height=0.52\columnwidth,
      ymin=0, ymax=100,
      xmin=0.70,
      xlabel={$\log_{10}$ token count},
      ylabel={Reconstruction accuracy (\%)},
      ymajorgrids,
      xmajorgrids,
      grid style={gray!15},
      tick label style={font=\scriptsize},
      label style={font=\scriptsize},
      legend style={
        at={(0.5,1.02)},
        anchor=south,
        draw=none,
        fill=none,
        font=\scriptsize,
        legend columns=3,
        /tikz/every even column/.append style={column sep=0.8em},
      },
      every axis plot/.append style={line width=0.7pt, mark=*, mark size=1.3pt},
    ]
      \addplot coordinates {
        (0.7688,29.2827) (0.9084,45.3429) (1.0480,61.2511) (1.1876,71.1823)
        (1.3272,79.7013) (1.4668,85.0250) (1.6064,90.0053) (1.7460,91.8871)
        (1.8856,93.5981) (2.0253,95.1463) (2.1649,96.6119) (2.3045,96.2461)
        (2.4441,97.5240) (2.5837,97.1310) (2.7233,97.0465) (2.8629,97.5492)
        (3.0025,97.1833) (3.1421,99.1716) (3.2817,99.0363) (3.4213,99.3479)
        (3.5610,99.1653) (3.7006,98.8730) (3.8402,99.2784) (3.9798,99.6517)
        (4.1194,99.9201) (4.2590,99.4327) (4.3986,99.9785) (4.5382,99.8011)
        (4.6778,99.8854) (4.8174,99.9854)
      };
      \addplot coordinates {
        (0.7688,43.7453) (0.9084,60.5202) (1.0480,74.3712) (1.1876,82.4053)
        (1.3272,88.3629) (1.4668,92.2385) (1.6064,95.6898) (1.7460,96.8922)
        (1.8856,97.7954) (2.0253,98.4220) (2.1649,99.3921) (2.3045,98.9176)
        (2.4441,99.8135) (2.5837,98.8719) (2.7233,98.9323) (2.8629,98.9476)
        (3.0025,98.2639) (3.1421,99.9907) (3.2817,99.9963) (3.4213,99.9985)
        (3.5610,99.9985) (3.7006,100.0000) (3.8402,100.0000) (3.9798,100.0000)
        (4.1194,100.0000) (4.2590,100.0000) (4.3986,100.0000) (4.5382,99.9972)
        (4.6778,100.0000) (4.8174,100.0000)
      };
      \addplot coordinates {
        (0.7688,48.6211) (0.9084,64.9858) (1.0480,77.4911) (1.1876,84.7229)
        (1.3272,89.8649) (1.4668,93.3528) (1.6064,96.3658) (1.7460,97.3944)
        (1.8856,98.2148) (2.0253,98.6743) (2.1649,99.5381) (2.3045,99.0556)
        (2.4441,99.9223) (2.5837,98.9048) (2.7233,98.9643) (2.8629,98.9719)
        (3.0025,98.2738) (3.1421,99.9969) (3.2817,99.9982) (3.4213,100.0000)
        (3.5610,99.9985) (3.7006,100.0000) (3.8402,100.0000) (3.9798,100.0000)
        (4.1194,100.0000) (4.2590,100.0000) (4.3986,100.0000) (4.5382,100.0000)
        (4.6778,100.0000) (4.8174,100.0000)
      };
      \legend{Top-1, Top-5, Top-10}
    \end{axis}
  \end{tikzpicture}
  \caption{Accuracy vs. token frequency on a 2M slice of OpenWebText.}
  \label{fig:freq_vs_acc}
\end{figure}

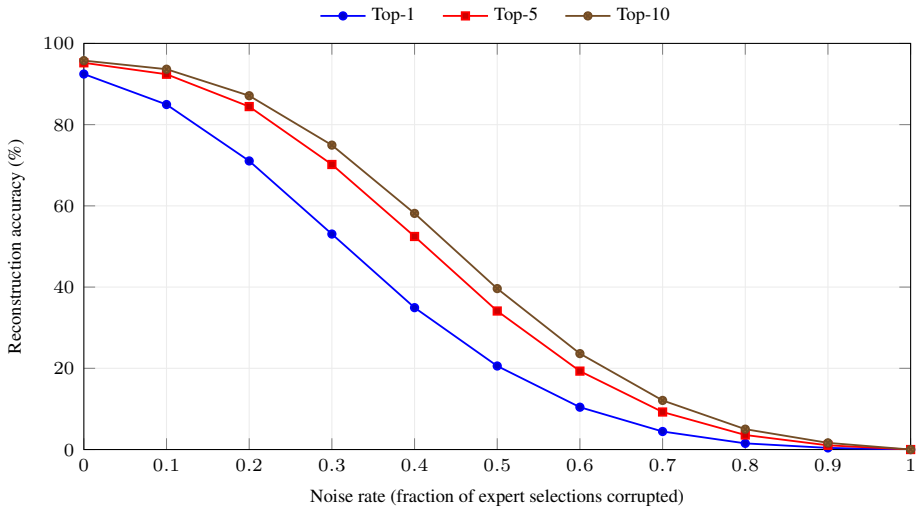
\begin{figure}[t]
  \centering
  \begin{tikzpicture}
    \begin{axis}[
      width=0.9\columnwidth,
      height=0.50\columnwidth,
      ymin=0, ymax=100,
      xmin=0, xmax=1.0,
      xlabel={Noise rate (fraction of expert selections corrupted)},
      ylabel={Reconstruction accuracy (\%)},
      ymajorgrids,
      xmajorgrids,
      grid style={gray!15},
      tick label style={font=\scriptsize},
      label style={font=\scriptsize},
      legend style={
        at={(0.5,1.02)},
        anchor=south,
        draw=none,
        fill=none,
        font=\scriptsize,
        legend columns=3,
        /tikz/every even column/.append style={column sep=0.8em},
      },
      every axis plot/.append style={line width=0.7pt, mark=*, mark size=1.4pt},
    ]
      \addplot coordinates {
        (0.0,92.46) (0.1,84.95) (0.2,71.06) (0.3,53.06) (0.4,34.935)
        (0.5,20.57) (0.6,10.425) (0.7,4.44) (0.8,1.52) (0.9,0.40)
        (1.0,0.0)
      };
      \addplot coordinates {
        (0.0,95.225) (0.1,92.405) (0.2,84.445) (0.3,70.185) (0.4,52.46)
        (0.5,34.11) (0.6,19.31) (0.7,9.24) (0.8,3.58) (0.9,1.035)
        (1.0,0.0)
      };
      \addplot coordinates {
        (0.0,95.77) (0.1,93.65) (0.2,87.12) (0.3,74.955) (0.4,58.145)
        (0.5,39.63) (0.6,23.62) (0.7,12.095) (0.8,5.015) (0.9,1.645)
        (1.0,0.0)
      };
      \legend{Top-1, Top-5, Top-10}
    \end{axis}
  \end{tikzpicture}
  \caption{Expert-selection noise vs. reconstruction accuracy on OpenWebText.}
  \label{fig:noise_sweep}
\end{figure}

\paragraph{Expert-noise robustness.}
To simulate measurement error and probabilistic or imperfect mitigations, we evaluate robustness to expert-selection trace noise by independently corrupting a fraction $p$ of the observed expert selections by replacing them with a uniformly random expert, and report top-$k$ token recovery accuracy as a function of $p$ (\Cref{fig:noise_sweep}).

For settings where expert selections may leak through side channels \citep{ding2025moecho}, we view several engineering mitigations as reasonable: (i) reduce the distinguishability of expert execution by balancing expert workloads and memory access patterns; (ii) add dummy compute or constant-work padding to blur expert-dependent activity; (iii) introduce randomness in routing (e.g., logit noise) or periodically permute expert identity to reduce trace stability; (iv) harden the routing/expert execution boundary (e.g., isolate co-resident workloads or disable exposure of fine-grained performance counters); and (v) make physical side channel measurements more difficult by shielding against leakage and removing or securing nearby sensors.

These defenses may incur performance or quality costs (e.g., increased compute or perturbed routing decisions); we leave a quantitative evaluation of the tradeoffs to future work.

\section{Limitations}
\paragraph{Scalability to long sequences.}
Our strongest results are for short sequences (32 tokens), and while the sequence decoder can recover useful information for longer windows, we have not systematically characterized the limits of inversion as sequence length grows (e.g., hundreds or thousands of tokens). Longer contexts increase ambiguity and may require more expressive architectures or search procedures.

\paragraph{Access and transferability.}
Our threat model assumes an adversary can obtain expert-selection traces and a compatible tokenizer, and that the adversary can collect sufficient ``(text, expert-selection trace)'' training pairs for a learning-based inverter (e.g., via an instrumented model instance, a same-family model, or internal logs). In practice, transfer to different model families, tokenizers, routing configurations, or expert permutations may reduce reconstruction quality; we do not evaluate cross-model transfer.

\paragraph{Partial traces.}
Although our threat model allows observing only a subset of layers, our main evaluations assume expert selections from all 24 layers. We evaluated information content per layer but have not measured reconstruction accuracy with such partial information since that requires retraining the model.

\paragraph{Decoder complexity.}
Our sequence decoder uses a learned encoder-only transformer and benefits from large-scale training data. More sophisticated decoding procedures (e.g., beam search or iterative refinement on top of the decoder) may further improve reconstruction, but we leave such extensions to future work.

\subsubsection*{Acknowledgments}
This work was supported by \href{https://www.matsprogram.org/}{\matslogo\ MATS} and \href{https://sparai.org/}{\sparlogo\ SPAR}. Various contributions were made by Jacob Lagerros, Natalia Kokoromyti, Luc Chartier, George Tourtellot and Krystal Maughan.

\section*{Ethics Statement}
This work studies privacy leakage mechanisms in MoE language model deployments. While it can inform defensive engineering (e.g., treating routing traces as sensitive outputs and hardening against side channels), it also could be misused to recover private user prompts when routing signals leak. We therefore emphasize mitigations and recommend minimizing exposure of routing traces and considering side-channel-resilient deployment practices for MoE inference. OpenWebText contains sensitive data; however, the original text is tokenized and isn't explicitly stored. By omitting details of methods to extract expert identities from side channel measurements, this work primarily advances the theoretical understanding of this type of attack rather than providing a practical method to breach confidentiality in real-world deployments.

\bibliographystyle{iclr2026_conference}
\bibliography{references}

\appendix
\section{Appendix}
\subsection{MoE Router Mechanism}
We briefly describe standard top-$k$ routing used in many MoE transformers. In OpenAI's reference implementation of \texttt{gpt-oss-20b}, routing occurs in the MoE MLP sublayer: given the MLP input vector $x \in \mathbb{R}^d$ (i.e., the residual stream entering the MoE MLP, after the attention sublayer), routing proceeds as follows:
\begin{enumerate}
  \item \textbf{Normalize.}
  \begin{equation}
    h = \mathrm{RMSNorm}(x).
  \end{equation}

  \item \textbf{Score each expert.} Compute logits over experts with a single affine map (where $n$ is the number of experts):
  \begin{equation}
    s = W_r h + b \in \mathbb{R}^n.
  \end{equation}
  We refer to $s$ as the \emph{expert logits}.

  \item \textbf{Pick the experts.} Select indices of the top-$k$ experts:
  \begin{equation}
    I = \mathrm{TopK}(s, k).
  \end{equation}
  We call $I$ the \emph{expert selections}.

  \item \textbf{Compute mixing weights.} Convert scores to nonnegative weights and normalize across the selected experts (e.g., using a softmax over $s_I$):
  \begin{equation}
    \alpha = \mathrm{softmax}(s_I).
  \end{equation}

  \item \textbf{Run selected experts.} Each selected expert $e_i$ processes $h$ (e.g., a SwiGLU MLP) to produce
  \begin{equation}
    y_i = e_i(h).
  \end{equation}

  \item \textbf{Combine.} Take the weighted sum:
  \begin{equation}
    y = \sum_{i \in I} \alpha_i\, y_i.
  \end{equation}

  \item \textbf{Residual add.} Return $x + y$ to the residual stream.
\end{enumerate}

In simple terms, each token gets a score for each expert (via a learned gate matrix and bias), is routed through the $k$ most preferred experts, and the resulting expert outputs are mixed and added back to the residual stream. In our main setting, OpenAI's \texttt{gpt-oss-20b} has 24 layers, 32 experts per layer, and routes top-4 experts per token.

\end{document}